\documentclass[conference]{IEEEtran}
\usepackage{times}
\usepackage{comment}

\usepackage[numbers]{natbib}
\usepackage{multicol}
\usepackage{graphicx}
\usepackage{subcaption}
\usepackage{amsmath,amsthm,amsfonts,amssymb,amscd}
\usepackage[bookmarks=true]{hyperref}

\usepackage[normalem]{ulem}
\usepackage[colorinlistoftodos]{todonotes}
\newcommand{\wenzhen}[1]{\todo[inline,color=red!40]{Wenzhen: #1}}
\newcommand{\joe}[1]{\todo[inline,color=blue!20]{Joe: #1}}
\newcommand{\yifan}[1]{\todo[inline,color=orange!20]{Yifan: #1}}

\begin{document}

\title{Understanding Dynamic Tactile Sensing for Liquid Property Estimation}


\author{\authorblockN{Hung-Jui Huang\hspace{60pt}    Xiaofeng Guo \hspace{60pt}   Wenzhen Yuan}
\vspace{7pt}
\authorblockA{Carnegie Mellon University \\
Email: \{hungjuih, xguo2, wenzheny\}@andrew.cmu.edu
}}


%

\maketitle

\begin{abstract}
Humans perceive the world by interacting with objects, which often happens in a dynamic way. For example, a human would shake a bottle to guess its content. However, it remains a challenge for robots to understand many dynamic signals during contact well. This paper investigates dynamic tactile sensing by tackling the task of estimating liquid properties. We propose a new way of thinking about dynamic tactile sensing: by building a light-weighted data-driven model based on the simplified physical principle. The liquid in a bottle will oscillate after a perturbation. We propose a simple physics-inspired model to explain this oscillation and use a high-resolution tactile sensor GelSight to sense it. Specifically, the viscosity and the height of the liquid determine the decay rate and frequency of the oscillation. We then train a Gaussian Process Regression model on a small amount of the real data to estimate the liquid properties. Experiments show that our model can classify three different liquids with 100\% accuracy. The model can estimate volume with high precision and even estimate the concentration of sugar-water solution. It is data-efficient and can easily generalize to other liquids and bottles. Our work posed a physically-inspired understanding of the correlation between dynamic tactile signals and the dynamic performance of the liquid. Our approach creates a good balance between simplicity, accuracy, and generality. It will help robots to better perceive liquids in different environments such as kitchens, food factories, and pharmaceutical factories.
\end{abstract}

\IEEEpeerreviewmaketitle

\section{Introduction}
Perceiving and understanding the physical world is a core capability of intelligent robots. Among the various sensing modalities, tactile sensing is typically used to understand physical interactions. For example, humans learn about an object's hardness by pressing it \cite{srinivasan1995} and 
a liquid's viscosity by shaking its container. They achieve these tasks using dynamic tactile sensing, which focuses more on changes of the signal across time. Dynamic tactile sensing is in contrast to static tactile sensing, which focuses on understanding the signal (force distribution, contact surface deformation) at one shot (e.g., localizing contact). Despite the importance of dynamic tactile sensing, its use for robots is not as well studied as it is for humans.

Among the common perceptual tasks requiring contacts, we study robot dynamic tactile sensing by tackling the task of noninvasively estimating the properties of liquid inside a container. Liquid properties are often closely related to their dynamics. For example, liquid viscosity refers to the restraining force when liquid flows, which requires us to focus on the temporal dimension of the dynamic tactile signal. Further, liquid property estimation has a wide robotics application. For example, monitoring viscosity online provides immediate feedback for operations in the food processing industry \cite{CULLEN2000451}.
However, it remains a challenge for robots to reliably estimate liquid properties without using specialized tools like viscometers.

\begin{figure}[t]
    \centering
    \includegraphics[width = 0.95\linewidth]{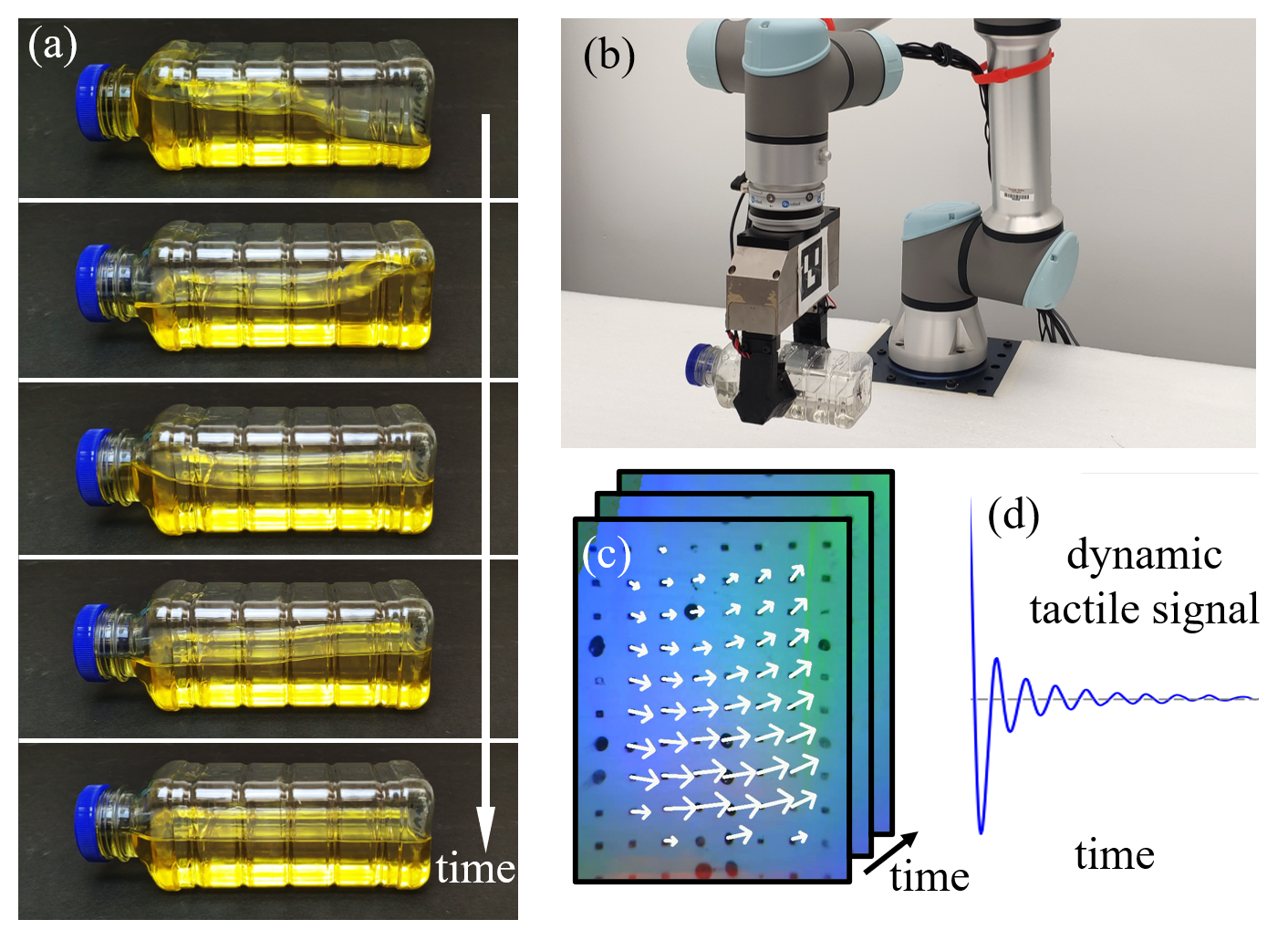}
    \caption{(a) The fluid in a bottle oscillates after a lateral perturbation. (b) We build a robot setup to shake containers. (c) Tactile signal from finger-tip GelSight sensors after the perturbation. (d) The principal motion of the GelSight markers after a perturbation on the bottle.}
    \label{fig:fig1}
\end{figure}


There are a number of works on content property estimation \cite{chen2016learning} \cite{8593838} \cite{matl} \cite{saal}. Inspired by humans, most approaches choose to shake the container and apply data-driven models to the received dynamic tactile signal. They show results on classifying $5$ to $12$ solids or fluids in a container with $90\%$ to $95\%$ accuracy using a variety of sensors. These works are limited to classifying objects with very different physical properties (e.g., water and glycerine). Physics-based approach with sophisticated fluid dynamics analysis \cite{matl} achieved $98\%$ accuracy but only work on cylinder-shaped containers with precisely-measured size. Both data-driven and physics-based approaches require that content weight and volume are either fixed \cite{guler} \cite{saal} or pre-estimated by other methods \cite{matl}. They also cannot generalize to different-shaped containers.

In this work, in addition to this content classification task, we tackle a much more challenging task: estimating the viscosity and liquid volume given a bottle of sugar-water solution. It requires distinguishing sugar-water solutions with different concentrations based on slight differences in viscosity, agnostic to the volume of liquid in the container.

We interpret the liquid-property-relevant tactile signal with a physics-inspired approach, which complements physics-based with data-driven modeling to leverage the strengths of both. By understanding how dynamic tactile signal is generated, we construct a model using simple physics. In addition, the complex, unmodeled effects of fluid dynamics in irregularly shaped containers are modeled by a data-driven method. Our method starts by approximating our liquid-container system as a second-order system based on our observation that the movement of the liquid surface after perturbation is a damped oscillation. Based on physics, We show that the oscillation frequency and decay rate of the damped oscillation are directly correlated with the liquid's height and viscosity. Therefore, the robot in our method first lifts, perturbs, and holds still the liquid container, and then collects the dynamic tactile signal that reflects the damped oscillation movement of the liquid. With our high-precision, vision-based touch sensor GelSight \cite{s17122762}, this damped oscillation movement is precisely detected after filtering, as shown in Fig. \ref{fig:fig1}d. We then learn a data-efficient model which takes the decay rate and the oscillation frequency of the dynamic tactile signal, and infers the liquid viscosity and volume (quantified as height) with very high precision.

\textbf{Contributions: }
Our main contribution is solving a dynamic tactile sensing task using an approach that maintains a good balance between physics-based and data-driven modeling.
We also contribute to estimating liquid properties with higher precision and greater generalizability. Our model achieves $100$\% classification accuracy on water, oil, and detergent. More importantly, our approach can estimate sugar concentration (ranging from $0$ to $160$ wt\%) with $15.3$ wt\% error and liquid height with $0.56$ mm error, using only tactile sensors. Note that $15.3$ wt\% sugar-water solution is less viscous than whole milk. Our height estimation is, in fact, so precise that the error may come from measuring the ground truth by rulers. Our method is data-efficient ($\leq 60$ training datapoints), agnostic to liquid volume, and can be easily generalized to different containers with similar shapes ($15$ fine-tuning datapoints). We believe approaches balancing physics-based and data-driven modeling can also be applied to other dynamic tactile sensing tasks (e.g., rubbing a surface to estimate its texture) since they often involve complex dynamics that cannot be purely modeled by physics or learned with limited real-world data.

\section{Related Work}

\subsection{Dynamic Tactile Sensing}

Dynamic tactile sensing is important in various robotics tasks related to perception, manipulation, locomotion, etc. \cite{de2009tactile} \cite{jamali2010material} \cite{kolvenbach2019haptic}. For example, \citet{sinapov2011vibrotactile} used a three-axis accelerometer to detect 20 different surfaces by performing 5 different exploratory scratching behaviors. They can recognize the surfaces with 80\% accuracy based on spectro-temporal features. \citet{taunyazov2021extended} used dynamic tactile sensing to achieve grasp stability prediction during object handover and accurate food identification through tools.
They also showed the ability to accurately localize taps on an acrylic rod based on vibrations caused by the tapping. \citet{giguere2011simple} used a tactile probe which is passively dragged along a surface for surface identification. They designed multiple features in 4-s time windows of dynamic tactile data and classified 10 different types of terrains. 

\subsection{Content Estimation}

Content estimation, or estimation of the object inside a given container, can be thought of as a specific type of object recognition. However, unlike other object recognition tasks, content estimation does not involve direct visual or physical access to the object, making the task much more challenging. 

There are multiple previous works utilizing different modalities to solve this task. For example, \citet{chen2016learning} used accelerometers and contact microphones to classify the mechanical parts in the container by shaking it.  \citet{guler} combined visual and tactile data to identify five types of contents. They used static deformation information of the paper container and achieved a 95\% classification accuracy. \citet{saal} used tactile data recorded during shaking to estimate the viscosity of different liquids with the same weight. \citet{matl} used the force-torque sensor to estimate liquid mass, volume, and viscosity. They achieved 98\% accuracy in classifying water, oil, and honey. \citet{jonetzko2020multimodal} integrated tactile and auditory data generated by shaking to classify 8 types of pills in 4 weight groups, achieving an average accuracy of 91\%. \citet{8593838} elicited auditory information by shaking to classify content types and predict the weight. 
\citet{jin2019open} used auditory information generated by rotation to classify 20 kinds of particles in the container. 

In previous works, researchers mainly focused on solid contents, which are easier to recognize. For example, \cite{chen2016learning} used different mechanical parts such as bearings and nuts; \cite{sinapov2014learning} used glass, rice, beans, and screws. Some works target liquids, but only study very different liquids, such as yogurt and water \cite{guler}, or water and glycerine \cite{saal}. In this work, in addition to classifying different liquids, we also target a much more challenging task: precise regression on liquid height and viscosity. 

In addition, most works adopted a purely data-driven approach and the learned model is specific to a certain group of weight and container. The only counter case we found is \cite{matl}, where the authors used an analytical model of liquid sloshing dynamics to estimate the properties of liquid in cylinder containers. However, the model requires the exact geometry of the container. In contrast, our physics-inspired model, combined with machine learning, is able to generalize to different liquids and containers and is agnostic to the amount of liquid. Besides, our approach is data-efficient and achieves better volume estimation and classification results on many shapes of containers without knowing their exact dimension.

\section{Dynamic Model of Liquid and Tactile Signal} \label{model}

\subsection{Overview} \label{overview}
We aim to estimate the viscosity and volume of the liquid in a bottle using only tactile sensors. Typically, solving this problem involves two steps: initiating the dynamics of the liquid-container by an activation motion, and estimating the liquid properties using the collected dynamic tactile signal. Our method studies, models, and parameterizes the interacting tactile force between the sensor and liquid-container. The correlation between the parameterized forces and the targeted liquid properties is further learned in a data-driven way.


Another key component of our approach is the way we activate the system. 
While shaking the liquid-container is the most popular activation motion, it causes complex, nonlinear dynamics like turbulence. Inspired by the importance of step response to system characterization, we command the robot to perform a ``unit step input": laterally perturbing the liquid-container and then holding it still. The key is to collect dynamic tactile signals only in the ``hold still" stage, where the system dynamics is much simpler and the signal is cleaner (Fig. \ref{fig:fig1}d). In the rest of this section, we explain and model the simple physics principles governing the received dynamic tactile signal collected this way.


\subsection{Linear Model for Liquid Oscillation} \label{lpm}

\begin{figure}[t]
\centering
\includegraphics[width=0.25\textwidth]{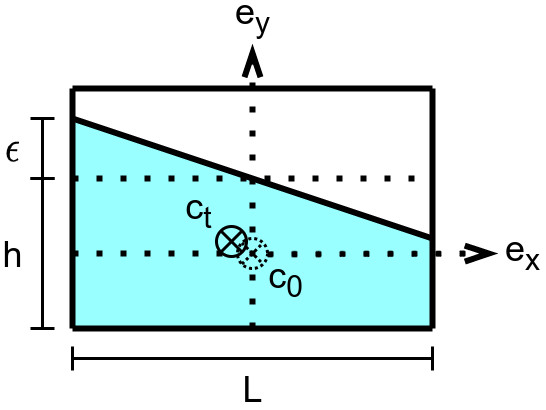}
\caption{Side view of liquid oscillating in a container. The center of mass of the liquid body is at $c_t$ in this perturbed state and $c_0$ in the steady-state.}
\label{fig:simple_mode}
\end{figure}

After perturbation, the liquid will oscillate in the container. The viscous force of the liquid, which dissipates energy from the system, causes the oscillation to decay. To model the oscillation, we consider a simplified system with liquid oscillating in the container in a simple mode, as shown in Fig. \ref{fig:simple_mode}. The length of the container is $L$, the height of the liquid is $h$, and the mass of the liquid is $m$. The entire container is actively held still in the air by a gripper with GelSight sensors attached. The origin of the system coordinate is set at the center of mass of the liquid body before oscillation. With the small offset $\epsilon(t)$ on the surface level from the steady-state, the center of mass of the liquid body is at $(-\epsilon L/6h, \epsilon^2/6h)$ and its velocity is $\approx -\dot{\epsilon} L /6h$. Assume the liquid's internal viscous force can be treated as an imaginary viscous force $F_\gamma(t)$ applied on the center of mass of the liquid body. We can write the principle of work and energy as:
\begin{equation}
\frac{d}{dt}\Big(\frac{1}{2} m (\frac{-\dot{\epsilon} L}{6h})^2 + mg\frac{\epsilon^2}{6h}\Big) = F_\gamma \cdot (\frac{-\dot{\epsilon} L}{6h}),
\label{eq_work_energy}
\end{equation}
The first and second terms are the kinetic and potential energy of the system, and the sum is the total energy. The RHS of Eq. \ref{eq_work_energy} is the energy dissipation rate caused by the viscous force. When the system perturbation $\epsilon(t)$ is small, the viscous force is approximately linear to the velocity \cite{physicalchemistry} of the center of mass and to the mass $m$: $F_\gamma = -\gamma m (\frac{-\dot{\epsilon} L}{6h})$. The constant factor $\gamma$ is determined by the liquid viscosity. For example, compared to water, oil has a stronger viscous force and larger $\gamma$, so the system energy dissipates faster. By replacing $F_\gamma$ in Eq. \ref{eq_work_energy} with $-\gamma m (\frac{-\dot{\epsilon} L}{6h})$, we get a linear second-order system:
\begin{equation}
\ddot{\epsilon}(t) + \gamma \dot{\epsilon}(t) + \frac{12gh}{L^2} \epsilon(t)=0.
\label{eq_linear}
\end{equation}
When the liquid is not too viscous ($\gamma$ is not too large), the solution of Eq. \ref{eq_linear} is a linear damped oscillation signal:
\begin{equation}
\epsilon(t) = Ae^{-\lambda t}\cos(\omega t + \phi)
\label{eq_solution},
\end{equation}
where the decay rate $\lambda = \frac{\gamma}{2}$ and the oscillating frequency $\omega = \sqrt{\frac{12gh}{L^2}-\frac{\gamma^2}{4}}$. The initial condition (the perturbation motion) determines the magnitude $A$ and the phase $\phi$. The solution matches our observation that after lateral perturbation, the liquid surface oscillates but gradually damps away.

As the center of mass of the liquid body oscillates, the robot gripper applies a lateral force $F_x$ on the side of the container to keep the container's position. The reaction force of $F_x$ is applied on the GelSight sensors, which record the force. The center of mass of the liquid body follows Newton's second law in the $x$ direction:
\begin{equation}
F_x = m a_x = m(-\ddot{\epsilon} L /6h).
\label{eq_newton}
\end{equation}
Taking the double derivative of the damped oscillation signal in Eq. \ref{eq_solution}, we have $\ddot{\epsilon}$ in Eq. \ref{eq_newton} being a similar signal with the same decay rate and oscillation frequency but a different magnitude and phase. Therefore, the dynamic tactile signal is:
\begin{equation}
F_x = Be^{-\lambda t}\cos(\omega t + \psi),
\label{eq_dts}
\end{equation}
which oscillates in the same pattern as the liquid surface $\epsilon$ in Eq. \ref{eq_solution} with a different magnitude $B$ and phase $\psi$. Eq. \ref{eq_dts} is our physics-inspired model and it explains the damped oscillating form of the dynamic tactile signal in Fig. \ref{fig:fig1}d.

Based on our model, the dynamic tactile signal has decay rate $\lambda = \frac{\gamma}{2}$, which directly relates to the liquid viscosity, and oscillation frequency $\omega = \sqrt{\frac{12gh}{L^2}-\frac{\gamma^2}{4}}$, which directly relates to the height of the liquid and liquid viscosity. It matches our observation that a more viscous liquid causes faster oscillation decay, and liquid with more volume oscillates at a higher frequency. In practice, our model, like many other physics models, is a simplification of the real world. For example, the decay rate $\lambda$ is non-linearly influenced by the shape of the container. We leave all the complex, unmodeled dynamics to be learned from the data-driven part of our approach (Section \ref{rlp}).

\subsection{Non-linear Energy Analysis} \label{nea}
Our approach estimates liquid viscosity and height using $\lambda$ and $\omega$ extracted from the dynamic tactile signal. However, extracting $\lambda$ and $\omega$ by fitting the signal using Eq. \ref{eq_dts} is not precise, because Eq. \ref{eq_dts} does not fit the signal well along the entire time range. As shown in Fig. \ref{fig:fig1}d, the signal decays faster at the beginning, not at the same rate throughout as modeled by Eq. \ref{eq_dts}. 
This is because the assumption that the viscous damping force $F_\gamma(t)$ is linear to the velocity of the center of mass does not hold when the system energy is high. In the linear case that $F_\gamma =  -\gamma m (\frac{-\dot{\epsilon} L}{6h})$, the energy dissipation rate $\gamma m (\frac{-\dot{\epsilon} L}{6h})^2$ is proportional to the system energy on average, so the system energy decays exponentially. When the system energy is high, the viscous damping force has higher order terms such as $F_\gamma = -\gamma m (\frac{-\dot{\epsilon} L}{6h}) - \kappa m (\frac{-\dot{\epsilon} L}{6h})^3$, and the energy dissipates faster than linear decay. Therefore, the system energy will decay faster than the exponential decay at the beginning, but behave approximately as an exponential decay once the system energy is lower. We take this affect into account when extracting $\lambda$ and $\omega$ from the dynamic tactile signal.

\section{Method}

\begin{figure*}[]
    \centering
    \includegraphics[width = 0.95\linewidth]{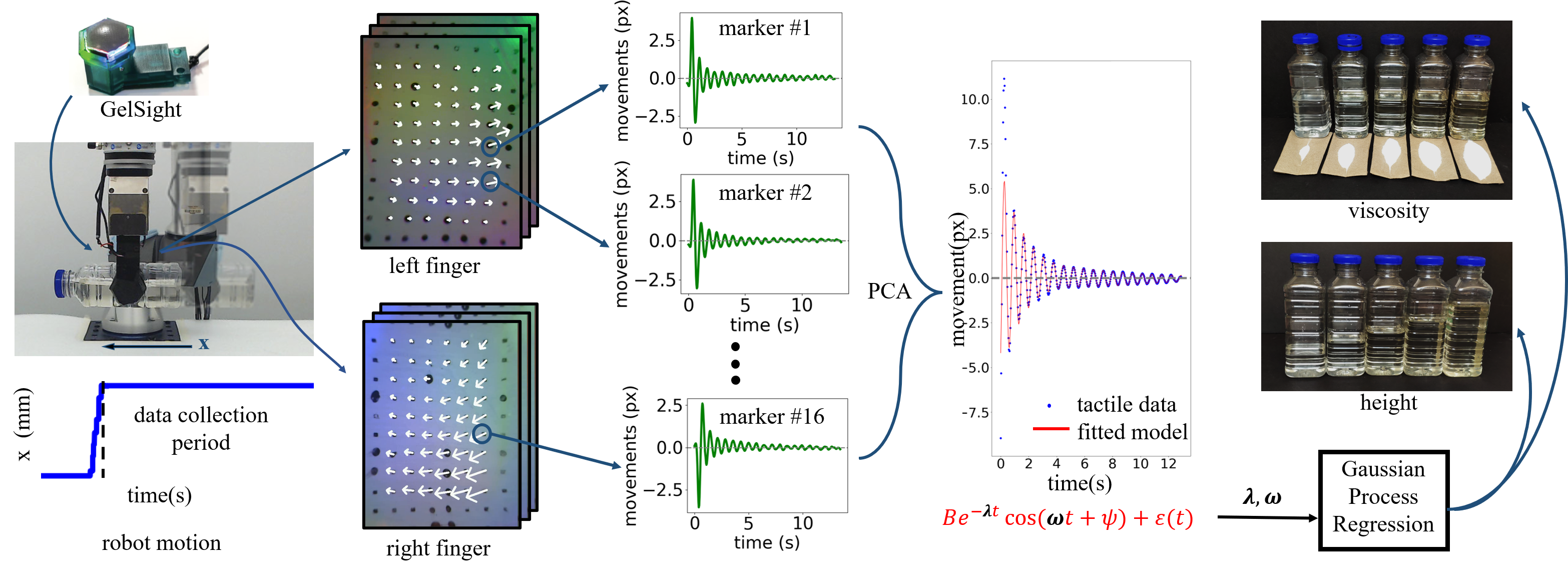}
    \caption{Pipeline of our method to estimate the viscosity and height of the liquid. We collect the tactile data during the free oscillation period after a lateral movement. We first track the marker motions from the GelSight \cite{s17122762} attached to both fingers. Then the top 16 markers with the largest motions are selected and a principal motion signal is calculated by taking PCA of their motions. We fit the principal motion to Eq. \ref{eq_target} and extract the oscillation frequency $\omega$ and decay rate $\lambda$ as the features. Then we fit a Gaussian Process Regression model taken $\omega$ and $\lambda$ as input to estimate the liquid height and viscosity.}
    \label{fig:pipeline}
\end{figure*}

In this section, we introduce our liquid properties estimation approach based on the physics-inspired liquid-container model presented in Section \ref{model}. Our approach first retrieves and de-noises the dynamic tactile signal that represents the lateral force $F_x$ from the GelSight images, and then extracts $\lambda$ and $\omega$ from Eq. \ref{eq_dts}. After that, we calculate the height of liquid $h$, liquid viscosity $\nu$, and other related physical properties of interest from the damping model with a data-driven model. The pipeline is shown in Fig. \ref{fig:pipeline}.

\subsection{Tactile Data Collection and Processing}
We use two GelSight sensors on a two-fingered gripper, one on each finger, to collect the dynamic tactile signal due to its high resolution. GelSight sensor is an vision-based tactile sensor consisting of a soft elastomer with printed markers, a lighting system, and an embedded camera. On contact with objects, the camera captures micron-level deformation of the sensing surface, which is an elastomer with markers \cite{s17122762}. It can estimate shear force distribution by tracking the markers' lateral motions in the image sequence collected by the embedded camera. When the liquid-container is held still after a lateral perturbation, we collect a sequence of GelSight images as the liquid oscillation motion subsides.

We extract marker contours in GelSight images by color segmentation and track their motion at sub-pixel accuracy. Since we are interested in signals in the low-frequency range, a $3$Hz low pass filter is applied on the marker motions for noise reduction. Marker motions are then ranked by their magnitude and only the top $16$ markers that are supposed to be in the center of contact are preserved. We concatenate the x, y motion of each marker to get a $32$-dimensional vector at each time step, and apply PCA to find the $32$-dimensional principal oscillating direction. In our case, the percentage of variance explained by the first principal component is above $90\%$. We project the $32$ dimensional vector at each time step to our principal oscillating direction and get our 1D de-noised dynamic tactile signal $u(t)$ that summarizes the damped oscillation motion of the top $16$ markers. This dynamic tactile signal $u$ is proportional to the shear force applied on GelSight, which is denoted by $F_x$ in Eq. \ref{eq_dts}.

\subsection{Fitting Linear Physics-Inspired Model}

Although the de-noised dynamic tactile signal $u$ should be in the form of linear damped oscillation (Eq. \ref{eq_dts}) in the ideal case, in practice, our signal $u$ is the sum of three main components: a linear damped oscillation $f$, a fast decay damped oscillation $g$, and a time-dependent offset $l$. As discussed in Section \ref{nea}, the dynamic tactile signal $u$ can be modeled as a linear damped oscillation $f$ as in Eq. \ref{eq_dts}, if we leave out the first few seconds where the system energy is high. However, the system starts with high energy, where the non-linear terms dominate the damping force and the signal decays faster. Therefore, we approximate $u$ with an additional term: a linear damped oscillation $g$ with a faster decay rate that only plays a part when the system energy is high. We also add an offset $l$ that is caused by in-hand slippage. $l$ is approximated as a quadratic function of time. In summary, we want to fit our dynamic tactile signal $u(t)$ with a parameterized function $\hat{u_\theta}(t)$:

\begin{equation}
\begin{aligned}
\hat{u_\theta}(t) &= \underbrace{Be^{-\lambda t}\cos(\omega t + \psi)}_{f} + \underbrace{B'e^{-\lambda' t}\cos(\omega' t + \psi')}_{g}\\
&+ \underbrace{c_2 (t)^2 + c_1 (t) + c_0,}_{l}
\end{aligned}
\label{eq_target}
\end{equation}
where $\theta=\{B,\lambda,\omega,\psi,B',\lambda',\omega',\psi',c_2,c_1,c_0\}$. Since $g$ by definition decays faster than $f$, the model is constrained by $\lambda'>\lambda$. To find $\theta$, we minimize a loss function $L(\theta)$ that defines an error metric between $\hat{u_\theta}(t)$ and $u(t)$. While $\theta$ in Eq. \ref{eq_target} consists of $11$ unknown parameters, only $\lambda$ and $\omega$ will be used to estimate the liquid properties.  
In order to better extract $\lambda$ and $\omega$, we care more about fitting the latter section of the signal well, where $f$ dominates the signal. Therefore, we choose a loss function that weighs the latter section of the signal more:

\begin{equation}
L(\theta) = \sum_{t}{\frac{(u(t) - \hat{u_\theta}(t))^2}{u(t)^2 + \delta}},
\label{eq_loss}
\end{equation}
where $\delta$ is a small value for numerical stability. We perform optimization to minimize the loss function $L(\theta)$ to get $\theta$. 

\subsection{Regression on Liquid Properties} \label{rlp}
Based on our physics-inspired model, the fitted parameters $\lambda$ and $\omega$ in $\theta$ are directly related to liquid physical properties: viscosity and height. We learn a model to approximate the complex dynamics that relates $\lambda$ and $\omega$ to the physical properties.
Using $\lambda$ and $\omega$ as features, we propose a parameterized quadratic regression model and a non-parameterized Gaussian Process Regression (GPR) model to estimate liquid height $h$ and viscosity $\nu$. Although one might consider estimating liquid height by measuring its weight using force-torque sensors, this approach requires a known liquid density, which can vary in a wide range (from $1 \text{ g}/\text{cm}^3$ to $1.3 \text{ g}/\text{cm}^3$ for sugar-water solutions used in our experiments).

Our analysis in Section \ref{lpm} suggests that $\lambda = \frac{\gamma}{2}$ and $\omega = \sqrt{\frac{12gh}{L^2}-\frac{\gamma^2}{4}}$, which means $h = \frac{L^2}{12g}(\omega^2 + \lambda^2)$. This motivates us to use quadratic models of $\lambda$ and $\omega$ to estimate both $h$ and $\mu = \log_{10}(\nu)$. Here, liquid viscosity $\nu$ is transformed to $\log_{10}$ space as in \cite{matl} and \cite{saal}. In practice, the irregular shape of the container and other unmodeled dynamics can cause some complex effects that cannot be fully captured by the quadratic model. We therefore propose to fit a GPR on $\lambda$ and $\omega$ to estimate both $h$ and $\mu$, which can capture those higher-order effects. 
In our experiments, we use GPR with a mixture of RBF kernel and white kernel. 
Thanks to the low-dimensional physics-inspired feature space $(\lambda, \omega)$, both the quadratic and GPR models predict liquid properties well without the need to use a more sophisticated model such as neural networks. In addition, our quadratic model has only $6$ parameters and our GPR model also requires very little data.

\subsection{Generalizing to Containers with Similar Shapes} \label{gcss}
A major advantage of using our physics-inspired features $(\lambda, \omega)$ is that we can generalize the property prediction models trained on one container to another container of similar shape, with only a few fine-tuning datapoints. Our experiments show that $h(\lambda, \omega)$ and $\mu(\lambda, \omega)$, as functions of $\lambda$ and $\omega$ in our trained model, have similar surface landscapes when trained with containers of similar shapes (as an example, compare Fig. \ref{fig:exp2_gp_c_pred} and Fig. \ref{fig:exp4_cuboid_pred}). To gain an intuitive understanding of this similarity, we can visit the relation $h = \frac{L^2}{12g}(\omega^2 + \lambda^2)$ in our physics-inspired model. When the size of the container $L$ changes, $h(\lambda, \omega)$ is simply scaled. We extend this observation to the estimation of viscosity $\mu$ and verify it in the experiments.

Assume we have container A and container B with similar shapes but different sizes. Given the trained model $\mu_A(\lambda, \omega)$ for container A, we parameterize the model in container B to have a similar landscape to $\mu_A(\lambda, \omega)$:
\begin{equation}
\mu_B(\lambda, \omega) = \mu_A(\alpha_1\lambda + \alpha_2, \beta_1\omega + \beta_2)
\label{eq_generalize}
\end{equation}
where $(\alpha_1, \alpha_2, \beta_1, \beta_2)$ are the shifting and scaling parameters. Instead of fitting a model for container B from scratch, we only need to fit the four parameters $(\alpha_1, \alpha_2, \beta_1, \beta_2)$, which require only a small amount of data.

\section{Experiments}
In the experiment section, we want to answer the following questions: (1) How well does our approach differentiate liquid contents that are similar to ones used in the related work \cite{matl} \cite{saal}? (2) How beneficial is it to use GelSight? (3) How well does our approach perform in the challenging liquid property regression task? (4) Can our approach generalize to other liquids or other containers?

\begin{figure}[t]
\centering
\includegraphics[width=0.95\linewidth]{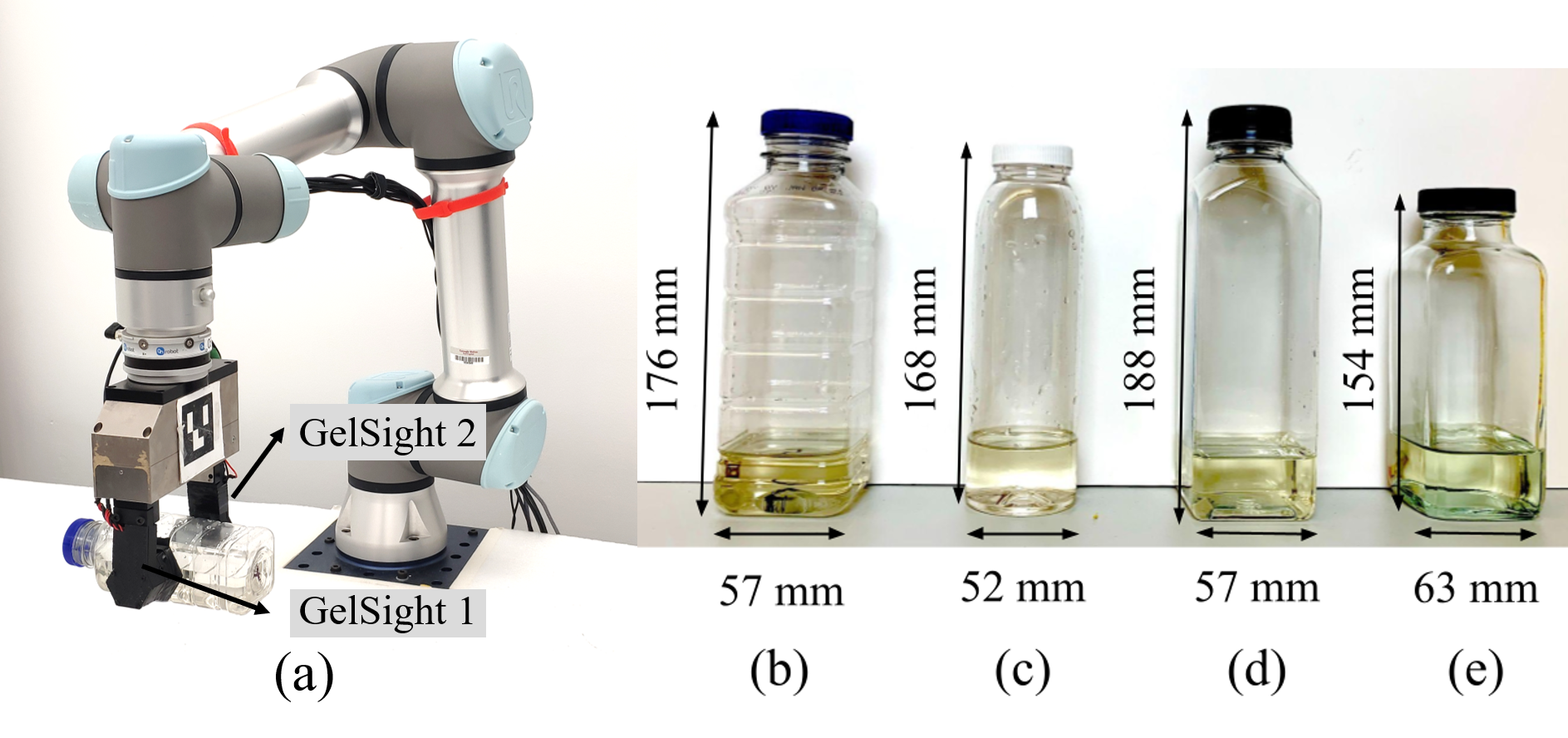}
\caption{(a) A 2-fingered gripper with two GelSights is used to grasp the bottle. (b) The grooved cuboid container. (c) The cylinder container.  (d) The cuboid container. (e) The glass container.}
\label{fig:containers}
\end{figure}

The experimental setup is shown in Fig. \ref{fig:containers}a. We use a $6$ DOF robot arm (UR5e by Universal Robotics) attached with a 2-fingered gripper (WSG50 by Weiss Robotics) to perform the activation motion. The dynamic tactile signal is captured by two Fingertip GelSight devices \cite{6943123} mounted on both fingers of the gripper. The elastomer on GelSight has markers printed in a $20\text{mm} \times 20\text{mm}$ region with an average distance of $1.8\text{mm}$ from each other. The camera in the sensor streams videos with $640 \times 480$ pixels resolution at $30 \text{Hz}$. We use a ubiquitous, grooved cuboid container (Fig. \ref{fig:containers}b) as the liquid container. For each trial, we manually place the container horizontally in the gripper. To prevent slippage, we place the container in the gripper at around the center of mass and cover the grasping region with tape to increase friction. We set the grasping force to be the largest force the gripper can apply, which is around 25N. After grasping, the robot arm lifts and activates the liquid-container with a short lateral movement. We then hold the system still and record the GelSight videos for another $13$ seconds. 


\subsection{Liquid Classification}
In this first experiment, we demonstrate our approach on a simple classification task. Our task is to classify three types of liquid with different viscosities: water ($1.1$ cSt), olive oil ($64$ cSt), and detergent ($440$ cSt). Ground-truth liquid viscosity is measured by a viscometer (NDJ-5S by JIAWANSHUN). In this task, liquid volume can vary anywhere from $1/3$ to $2/3$ full. We collect a training dataset of $24$ datapoints ($8$ different volumes for each liquid type, uniformly spanned in the range) and a testing dataset of $210$ datapoints ($70$ different volumes for each liquid type, randomly sampled in the range).

\begin{figure}[t]
\centering
\includegraphics[width=0.48\textwidth]{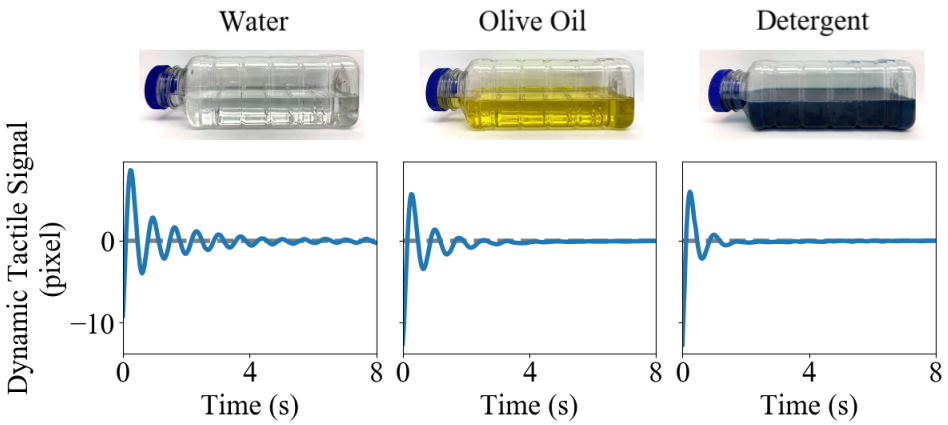}
\caption{Dynamic tactile signal $u$ when the container is half-filled with water, olive oil, and detergent, respectively.}
\label{fig:exp1_comparison}
\end{figure}

\begin{figure}[!tbp]
  \centering
  \hfill
  \begin{minipage}[b]{0.48\linewidth}
    \centering
    \subcaptionbox{\label{fig:exp1_pred}}
      {\includegraphics[width=\textwidth]{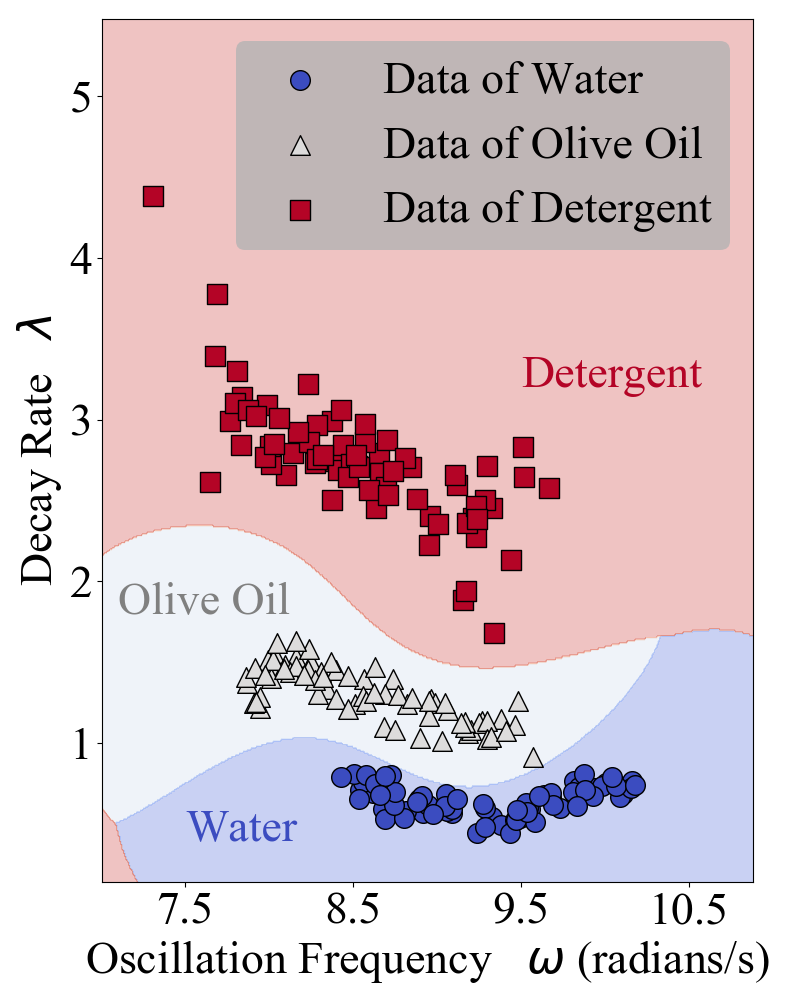}}
  \end{minipage}
  \hfill
  \begin{minipage}[b]{0.48\linewidth}
   \centering
   \subcaptionbox{\label{fig:exp1_height_frequency}}
    {\includegraphics[width=\textwidth]{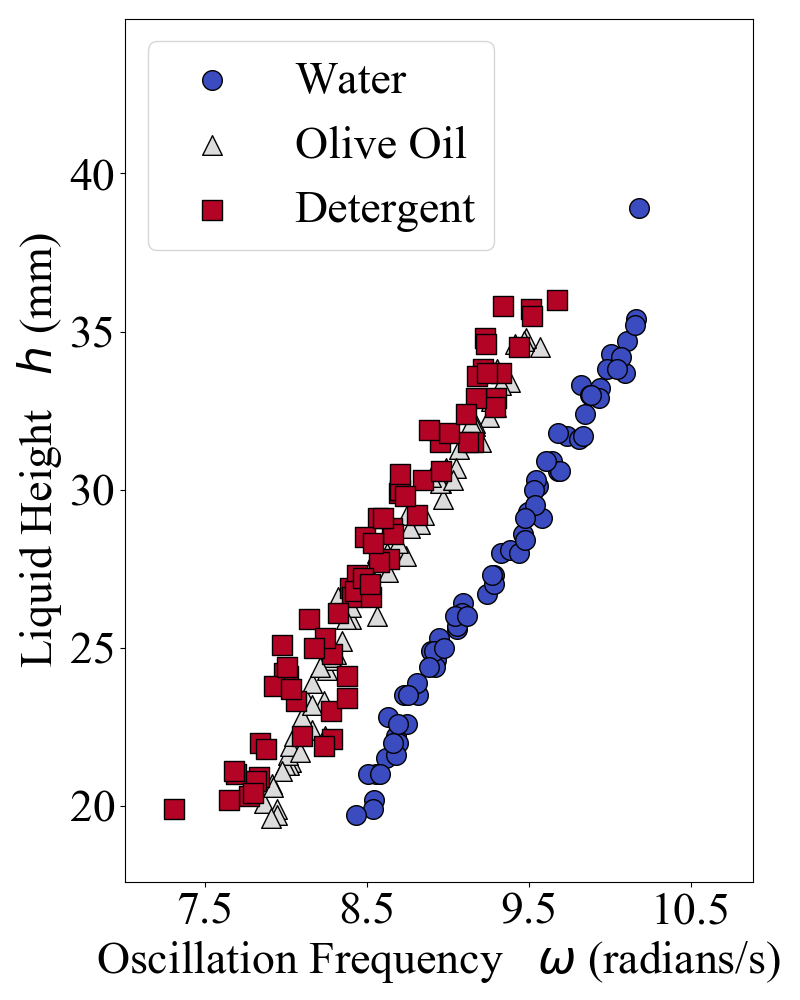}}
  \end{minipage}
  \hfill
  \caption{(a) Classification of water, olive oil, and detergent on the testing dataset. Colored regions represent the classification regions determined by the trained SVM. (b) Ocsillation frequency $\omega$ vs. liquid height $h$ for different liquid types.}
  \label{fig:exp1}
\end{figure}

Fig. \ref{fig:exp1_comparison} shows examples of dynamic tactile signals $u$ received with the three types of liquids. The dynamic tactile signal is recorded only in the ``hold still" stage after perturbation and is derived from GelSight marker motions. The signal decays fastest in detergent and slowest in water. Using the training dataset, we train an SVM classifier with an RBF kernel that takes $\lambda$ and $\omega$ as input. The distribution of the testing dataset and the decision boundaries of the trained classifier are shown in Fig. \ref{fig:exp1_pred}. Our classifier successfully classifies the testing dataset with an $100\%$ accuracy, which outperforms the $98\% \pm 10\%$ accuracy reported in \cite{matl}. The data in Fig. \ref{fig:exp1_pred} is well divided into $3$ bands corresponding to the three types of liquid. Within each band, we note that datapoints with lower $\omega$ are associated with a lower height $h$ (Fig. \ref{fig:exp1_height_frequency}).

\subsection{Comparison with Force-Torque Sensor Measurement} \label{force_torque}

\begin{figure}[t]
\centering
\includegraphics[width=0.48\textwidth]{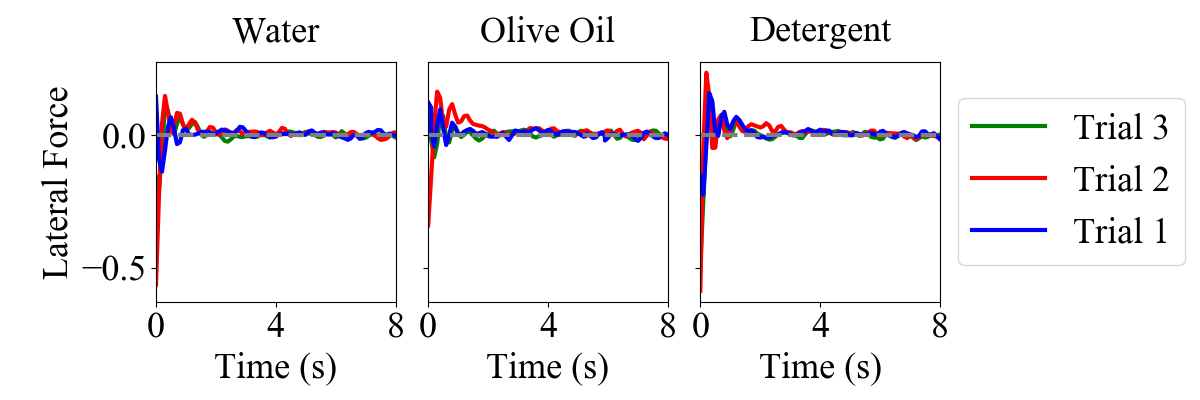}
\caption{Filtered lateral force measurement using a 6-axis force-torque sensor on the robot wrist after a perturbation. Data is collected with the same setup as in Fig. \ref{fig:exp1_comparison}. We achieve $34.3\%$ classification accuracy using this signal, which is only slightly better than the $33.3\%$ accuracy when guessing randomly.}
\label{fig:exp1_force_comparison}
\end{figure}

Since we take the lateral reaction force applied on GelSight as the input tactile signal, it comes naturally that one can use force-torque sensors to directly measure the force as in \cite{matl}. We mount a $6$-axis force-torque sensor (HEX-E by OnRobot) on the wrist of the robot arm to measure the dynamic tactile signal. Fig. \ref{fig:exp1_force_comparison} shows the filtered lateral force in $3$ repeated trials. While using a sensor with a similar specification sheet as in \cite{matl}, our force-torque sensor measurement has a much lower signal-to-noise ratio (SNR) and only achieves a classification accuracy of $34.3\%$ using the method presented in \cite{matl}. This difference might come from that our container being smaller with a less regular shape than \cite{matl} or the different way we perturb the system. In contrast, the GelSight signals have a much higher SNR and achieve $100\%$ accuracy. This experiment underlines the benefit of using GelSight.

\subsection{Estimating Sugar Concentration and Liquid Height} \label{sugar_water}
\begin{figure}[!tbp]
  \centering
  \begin{minipage}[b]{0.4\linewidth}
   \subcaptionbox{\label{fig:exp2_train_dataset}}
    {\includegraphics[width=\textwidth]{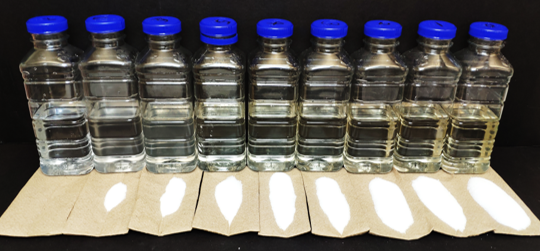}}
   \subcaptionbox{\label{fig:exp2_test_dataset}}
    {\includegraphics[width=\textwidth]{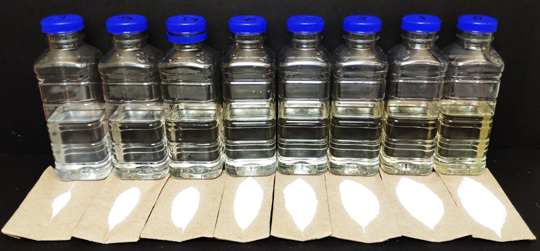}}
  \end{minipage}
  \hfill
  \begin{minipage}[b]{0.57\linewidth}
    \subcaptionbox{\label{fig:exp2_viscosity_concentration}}
      {\includegraphics[width=\textwidth]{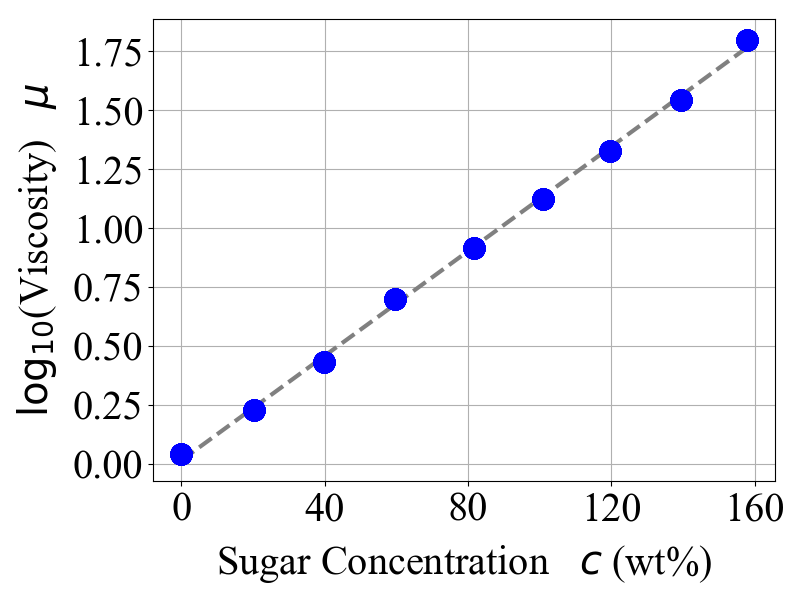}}
  \end{minipage}
  \caption{(a) Sugar-water solutions in the training dataset, concentration from $0$ wt\% (left) to $160$ wt\% (right)  (b) Sugar-water solutions in the testing dataset, concentration from $10$ wt\% (left) to $150$ wt\% (right)  (c) The relationship of $\log_{10}$ of viscosity and concentration of the sugar-water solutions.}
  \label{fig:exp2_dataset}
\end{figure}

We show the capability of our approach by demonstrating on a challenging experiment for liquid properties regression. We task our robot to estimate the concentration $c$, $\log_{10}$ of viscosity $\mu$, and volume (quantified as liquid height $h$) of sugar-water solutions. The concentration vary from $0$ to $160$ wt\%, making the viscosity vary from $1.1$ to $62.6$ cSt. We estimate viscosity in $\log_{10}$ scale as in \cite{matl} and \cite{saal} because liquid viscosity is an exponential function of sugar concentration. Fig. \ref{fig:exp2_viscosity_concentration} shows the relation between concentration and $\log_{10}$ of viscosity. In this experiment, the height of the liquid varies from $16$ mm (25\% full) to $40$ mm (75\% full). Out of this range, the liquid oscillation is too weak for perception. We make sugar-water solutions of $9$ different concentration levels equally spanned from $0$ to $160$ wt\% (Fig. \ref{fig:exp2_train_dataset}). For each concentration level, we collect data with $12$ different height levels equally spanned from $16$ to $40$ mm. Together, it forms the training dataset of size $108$.

The testing dataset is made of another set of sugar-water solution of $8$ different concentration levels equally spanned from $10$ to $150$ wt\% (Fig. \ref{fig:exp2_test_dataset}). For each solution, we setup the liquid-container system on the $12$ discretized liquid heights, same as the training dataset. We thus have $96$ testing setups. For each testing setup, we perform three independent tests. We then take the average $\lambda$ and $\omega$ of the $3$ signals from the same setup to reduce variance. We use this testing method for the following experiments as well.

\begin{figure}[!tbp]
  \centering
  \begin{minipage}[b]{0.98\linewidth}
   \subcaptionbox{\label{fig:exp2_concentration_comparison}}
    {\includegraphics[width=\textwidth]{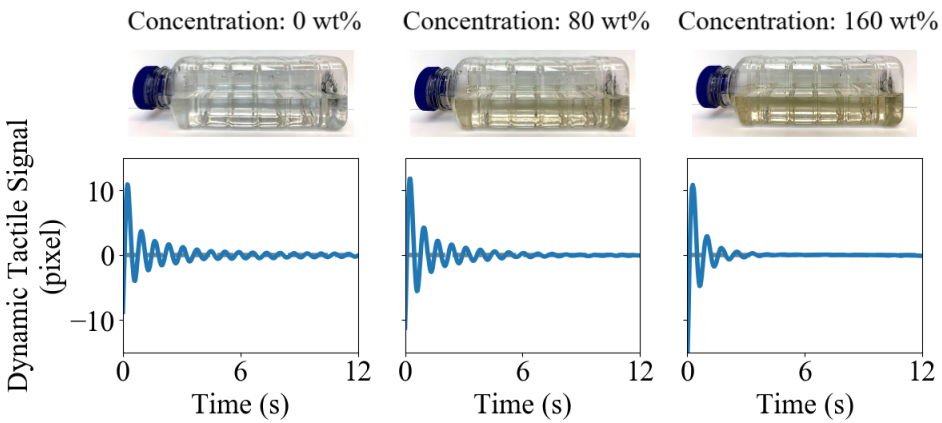}}
  \end{minipage}
  \hfill
  \begin{minipage}[b]{0.98\linewidth}
    \subcaptionbox{\label{fig:exp2_height_comparison}}
      {\includegraphics[width=\textwidth]{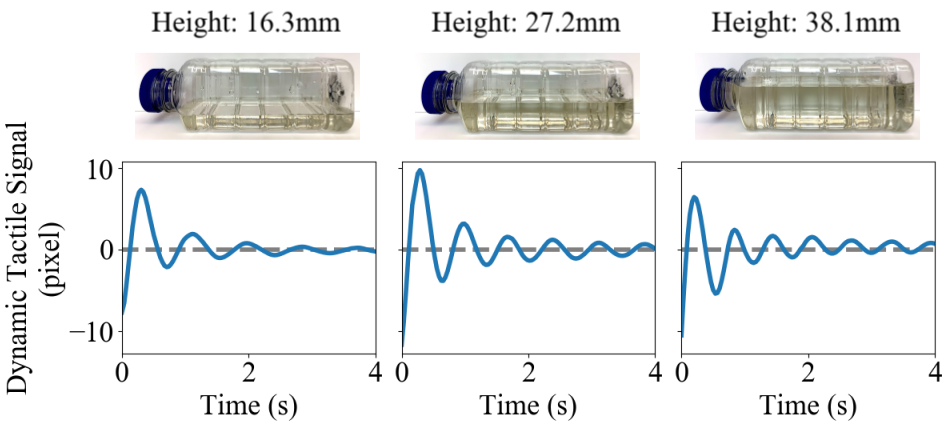}}
  \end{minipage}
  \caption{(a) Dynamic tactile signal of half-filled containers with sugar-water solutions of different concentration. (b) Dynamic tactile signal of containers with $70$ wt\% sugar-water solution at different liquid heights.}
  \label{fig:exp2_comparison}
\end{figure}

\begin{figure}[!tbp]
  \centering
  \begin{minipage}[b]{0.48\linewidth}
    \subcaptionbox{\label{fig:exp2_gp_h_pred}}
      {\includegraphics[width=\textwidth]{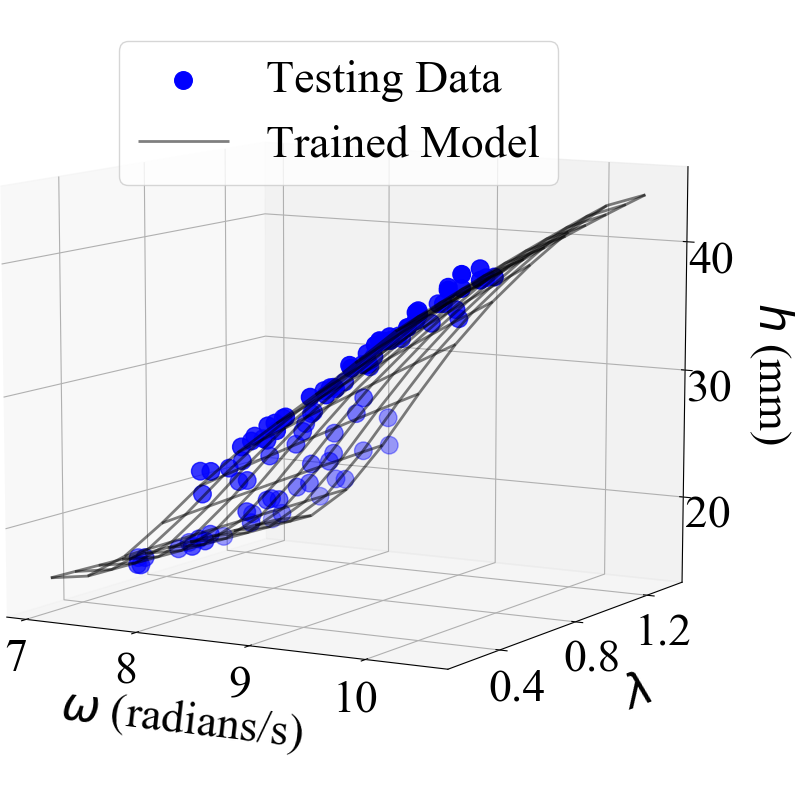}}
  \end{minipage}
  \hfill
  \begin{minipage}[b]{0.48\linewidth}
   \subcaptionbox{\label{fig:exp2_gp_c_pred}}
    {\includegraphics[width=\textwidth]{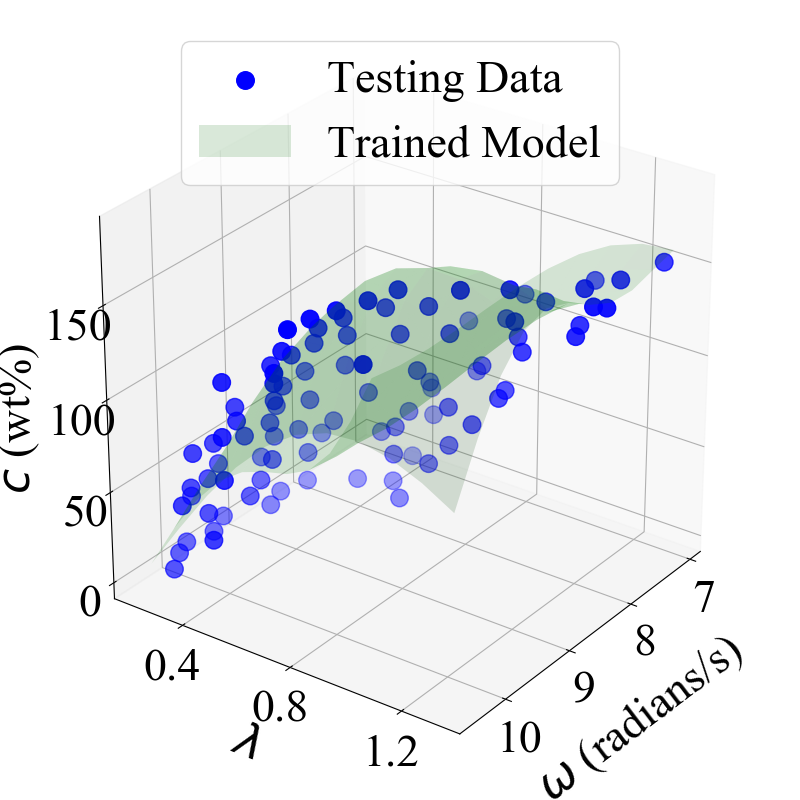}}
  \end{minipage}
  \hfill
  \begin{minipage}[b]{0.8\linewidth}
  \hfill
   \subcaptionbox{\label{fig:exp2_gp_h_conf}}
    {\includegraphics[width=0.48\textwidth]{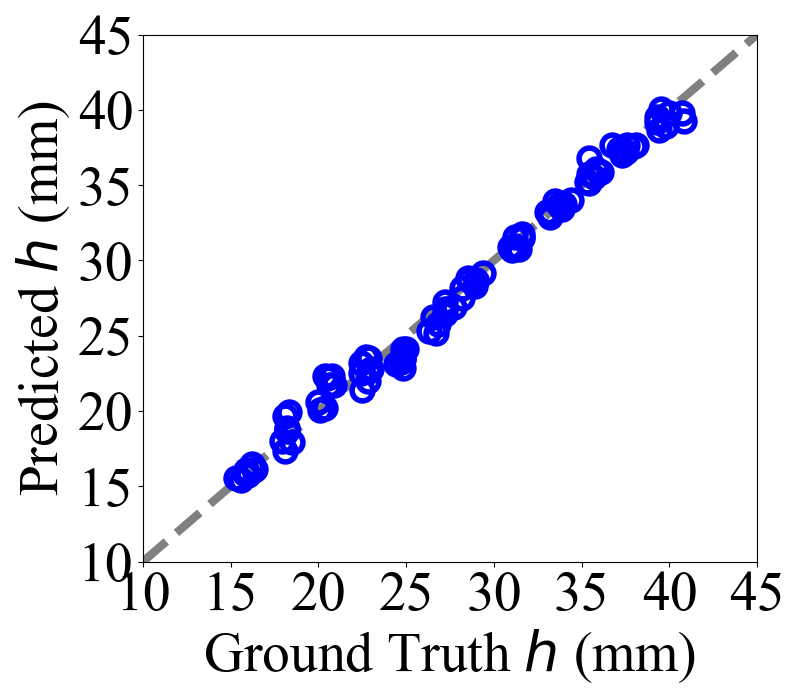}}
   \hfill
   \subcaptionbox{\label{fig:exp2_gp_c_conf}}
    {\includegraphics[width=0.48\textwidth]{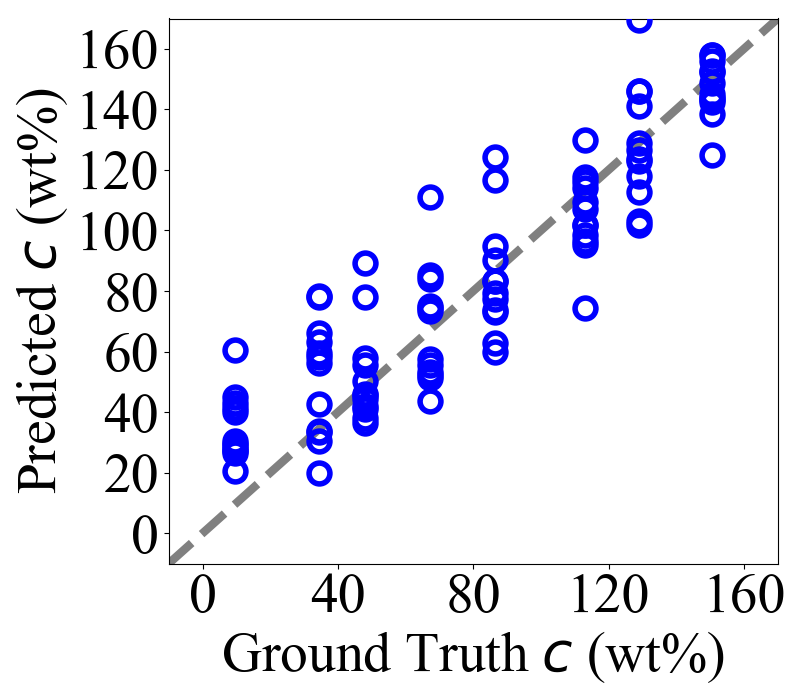}}
   \hfill
  \end{minipage}
  \caption{Prediction of liquid height and sugar concentration on the testing dataset using the GPR model. (a) Relation between liquid height $h$ and decay rate $\lambda$, oscillation frequency $\omega$ (b) Relation between sugar concentration $c$ and decay rate $\lambda$, oscillation frequency $\omega$ (c) Prediction of liquid height. (d) Prediction of sugar concentration.}
  \label{fig:exp2_gp}
\end{figure}

Fig. \ref{fig:exp2_concentration_comparison} shows examples of dynamic tactile signals when activating solutions of different concentration levels. Higher concentration levels result in faster signal decay. We also show the tactile signal for liquid at different heights in Fig. \ref{fig:exp2_height_comparison}. The signal oscillates faster when there is more liquid is in the container. For quantitative measurement of the volume and viscosity, we train three GPR models taking $\lambda$ and $\omega$ as input to predict $h$, $c$, and $\mu$, as introduced in Section \ref{rlp}. The fitting errors on the training and testing datasets are shown in Table \ref{tab:exp2_gp}. The prediction result on the testing dataset is shown in Fig. \ref{fig:exp2_gp}. We omit the prediction result of $\mu$ from Fig. \ref{fig:exp2_gp} because $\mu$ is almost linear to $c$ and their results have almost the same shape. The result shows that the testing data is close to the fitted GPR plane, and this model can estimate $h$, $c$, and $\mu$ with MAE of $0.56$ mm, $15.3$ wt\%, and $0.16$. Note that the ground truth of the liquid height has a measurement error of roughly $1.0$ mm. Results using GPR models also outperform results using parameterized quadratic models shown in the Appendix. In the rest of the paper, all results will refer to results on the testing dataset.

\begin{table}[ht]
\centering
\begin{tabular}{|c|c|c|c|}  
\hline 
 & $h$ (mm) & $c$ (wt\%) & $\mu$ \\ 
\hline 
\hline  
\hline
GPR Model MAE (training dataset) & 0.81 & 15.7 & 0.16 \\ 
\hline 
GPR Model MAE (testing dataset) & 0.56 & 15.3 & 0.16 \\ 
\hline 
\end{tabular}
\caption{Prediction error of liquid height $h$, sugar concentration $c$, and $\log_{10}$ of viscosity $\mu$ using the GPR models.}
\label{tab:exp2_gp}
\end{table}

\subsection{Generalization to Other Liquids}
Our models for predicting viscosity and liquid height trained on the sugar-water solutions can directly generalize to other liquids without re-training or fine-tuning. Using the GPR models trained in Section \ref{sugar_water}, we show results of estimating $\mu$ and $h$ of olive oil ($\mu = 1.81$), corn oil ($\mu = 1.72$), and mirin ($\mu = 1.34$). The testing dataset consists of $36$ setup in total: the liquid-container system on the $12$ discretized liquid height for the three types of liquid.

\begin{figure}[!tbp]
  \centering
  \hfill
  \begin{minipage}[b]{0.48\linewidth}
    \centering
    \subcaptionbox{\label{fig:exp3_height_pred}}
      {\includegraphics[width=\textwidth]{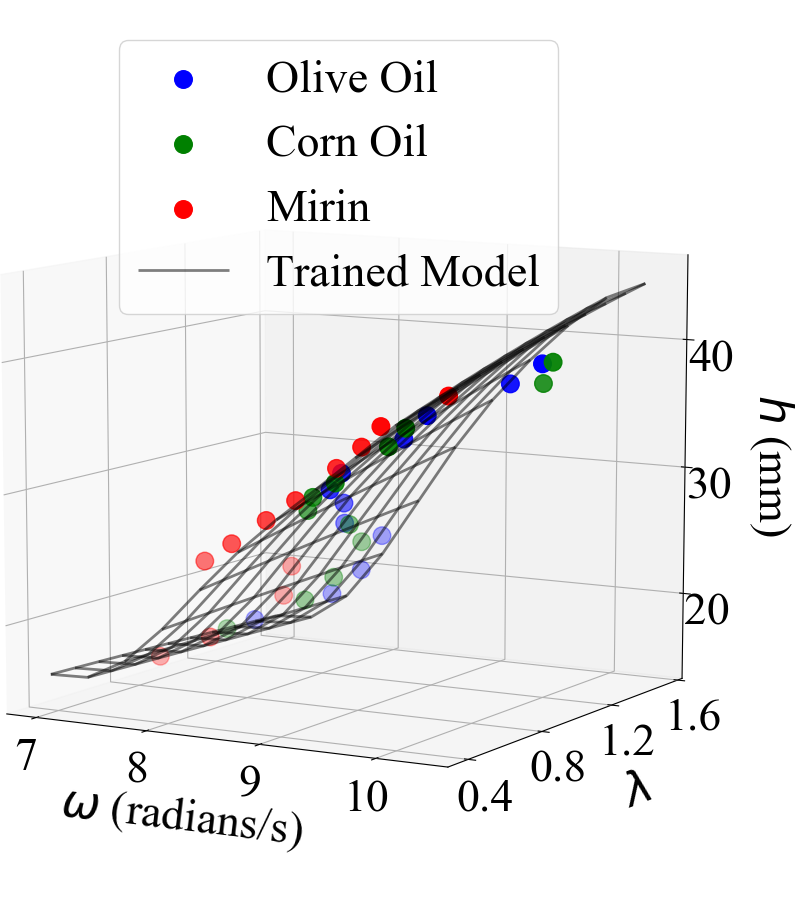}}
  \end{minipage}
  \hfill
  \begin{minipage}[b]{0.48\linewidth}
   \centering
   \subcaptionbox{\label{fig:exp3_mu_pred}}
    {\includegraphics[width=\textwidth]{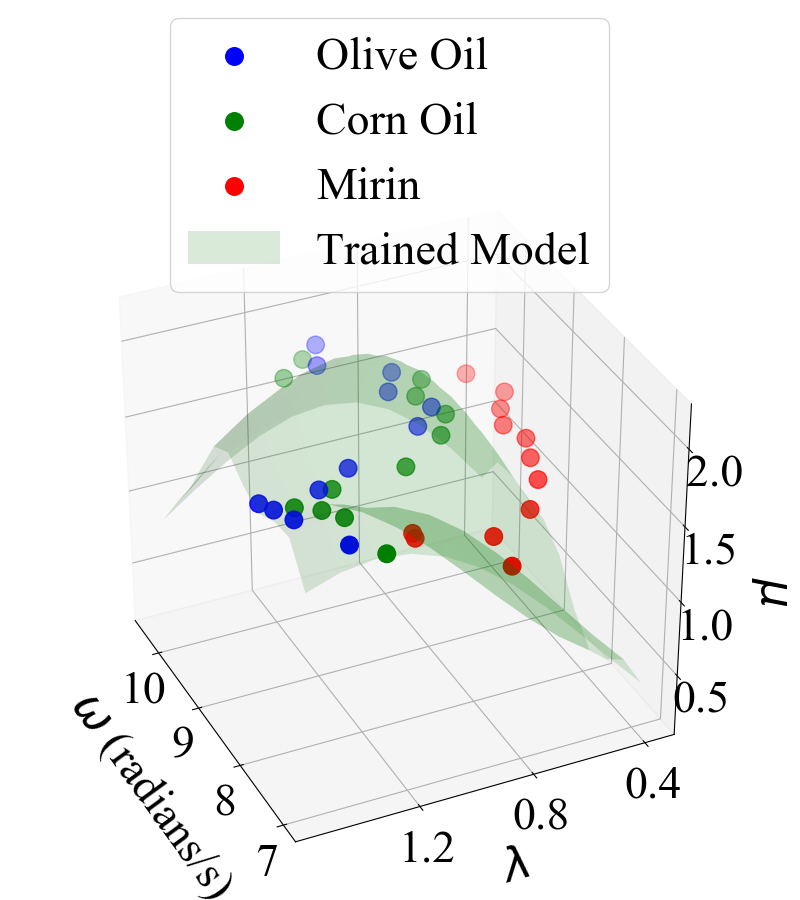}}
  \end{minipage}
  \hfill
  \caption{(a) The prediction result of liquid height $h$  on common liquids. (b) The prediction result of viscosity $\mu$ on common liquids (shown in $\log_{10}$ scale).}
  \label{fig:exp3}
\end{figure}

The prediction result is shown in Fig. \ref{fig:exp3} and Table \ref{tab:exp3}. The prediction error of $\mu$ in mirin is much more than that of the oils. We suspect that mirin might be a non-Newtonian fluid and its thickness can't be explained only by viscosity. \citet{saal} reported their mean square error (MSE) of $\mu$ being $0.48$. To compare, our MSE of $\mu$ is $0.02$ on olive oil, $0.03$ on corn oil, and $0.18$ on mirin.

\begin{table}[ht]
\centering
\begin{tabular}{|c|c|c|}  
\hline 
 & $h$ MAE (mm) & $\mu$ MAE \\ 
\hline 
\hline 
Olive Oil ($\mu = 1.81$)& 1.50 & 0.12 \\ 
\hline 
Corn Oil ($\mu = 1.72$)& 1.46 & 0.12 \\ 
\hline
Mirin ($\mu = 1.34$)& 0.99 & 0.35 \\ 
\hline 
\end{tabular}
\caption{Prediction error of liquid height $h$ and $\log_{10}$ of viscosity $\mu$ on common liquids.}
\label{tab:exp3}
\end{table}

\subsection{Experiments using Different Containers} \label{different_containers}
We show in this experiment that our approach works on containers of different sizes, shapes, and materials. Similar to Section \ref{sugar_water}, we collect training and testing datasets of sugar-water solutions (Fig. \ref{fig:exp2_train_dataset}, \ref{fig:exp2_test_dataset}) with different liquid heights using the three containers: cylinder container (Fig. \ref{fig:containers}c), cuboid container (Fig. \ref{fig:containers}d), and glass container (Fig. \ref{fig:containers}e). The cylinder container and cuboid container are made of plastic. To prevent slippage, we 3D-print a mold that clamps on the cylinder container and creates a flat contact surface for the gripper. The details of the datasets are shown in Table \ref{tab:exp4_dataset}. The cuboid container and the grooved cuboid container (Fig. \ref{fig:containers}b) used in the previous experiments have a similar shape but different sizes. The glass container is much heavier and is of a different shape and size than other containers.

\begin{table}[ht]
\centering
\begin{tabular}{|c|c|c|c|}  
\hline 
Container Type & Cylinder & Cuboid & Glass \\ 
\hline 
\hline 
Range of Level (mm)& 12 to 35 & 15 to 38 & 22 to 39 \\ 
\hline 
Trainig Dataset Size& 117 & 108 & 63 \\ 
\hline
Testing Dataset Size& 104 & 96 & 56 \\ 
\hline 
\end{tabular}
\caption{Details of the training and testing datasets collected on the cylinder, cuboid, and glass containers.}
\label{tab:exp4_dataset}
\end{table}

\begin{figure*}[]
    \centering
  \hfill
  \begin{minipage}[b]{0.27\linewidth}
    \centering
    \subcaptionbox{\label{fig:exp4_cylinder_pred}}
      {\includegraphics[width=\textwidth]{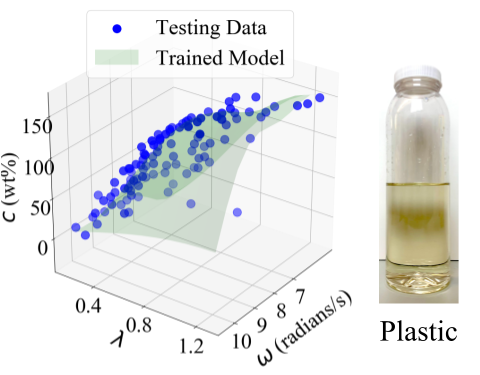}}
  \end{minipage}
  \hfill
  \begin{minipage}[b]{0.27\linewidth}
   \centering
   \subcaptionbox{\label{fig:exp4_cuboid_pred}}
    {\includegraphics[width=\textwidth]{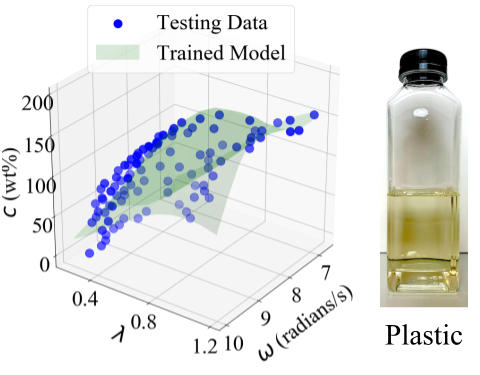}}
  \end{minipage}
  \hfill
  \begin{minipage}[b]{0.27\linewidth}
    \centering
    \subcaptionbox{\label{fig:exp4_cylinder_pred}}
      {\includegraphics[width=\textwidth]{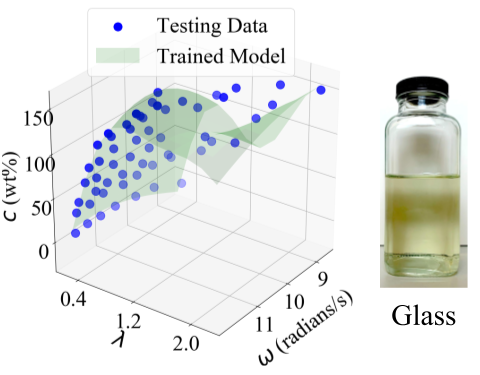}}
  \end{minipage}
  \hfill
  \caption{The sugar concentration $c$ prediction results with different containers. (a) Train and test with the cylinder container. (b) Train and test with the cuboid container. (c) Train and test with the glass container.}
  \label{fig:exp4}
\end{figure*}

The prediction results are shown in Fig. \ref{fig:exp4} and Table \ref{tab:exp4_results}. We can see that the trained GPR model using the grooved cuboid container (Fig. \ref{fig:exp2_gp_c_pred}) has a similar shape with the one of the cuboid container (Fig. \ref{fig:exp4_cuboid_pred}) but  different from the one of the cylinder container (Fig. \ref{fig:exp4_cylinder_pred}). This supports our discussion in Section \ref{gcss}. We also observe the GPR model having a very different landscape when using the glass container. A potential reason is that the liquid-container dynamics is different when the container weight is not negligible (the glass container has a similar weight to the liquid inside on average).

\begin{table}[ht]
\centering
\begin{tabular}{|c|c|c|c|}  
\hline 
 & $h$ (mm) & $c$ (wt\%) & $\mu$\\ 
\hline 
\hline 
MAE (Cylinder Container) & 0.84 & 13.5 & 0.15 \\ 
\hline 
MAE (Cuboid Container) & 1.96 & 20.4 & 0.23 \\ 
\hline 
MAE (Glass Container) & 0.91 & 12.9 & 0.14 \\ 
\hline 
\end{tabular}
\caption{Prediction error of liquid height $h$, sugar concentration $c$, and $\log_{10}$ of viscosity $\mu$ using GPR models in the cylinder, cuboid, and glass containers.}
\label{tab:exp4_results}
\end{table}

\subsection{Training Data Efficiency} \label{data_efficiency}
\begin{figure}[t]
\centering
\includegraphics[width=0.35\textwidth]{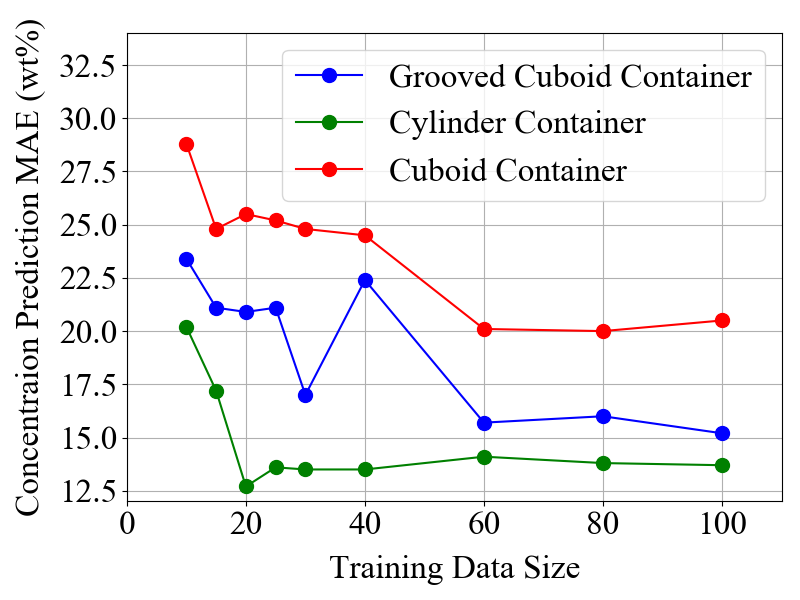}
\caption{The prediction error of sugar concentration when trained with different amount of data.}
\label{fig:exp5_data_efficiency}
\end{figure}

A key advantage of our approach is its data efficiency. We train GPR models using subsets of the training datasets in all three containers. The subsets have a size ranging from $10$ to $100$ data and are uniformly sampled from the original training datasets in Section \ref{sugar_water} and \ref{different_containers}. We only show results on concentration prediction because it is more challenging than liquid height prediction. The change of concentration prediction error when using different amounts of training data is shown in Fig. \ref{fig:exp5_data_efficiency}. For a smooth container like the cylinder container, we can train our GPR model with as little as $10$ data and has good prediction results ($\text{MAE} \leq 20 \text{ wt\%}$). The cuboid containers take $60$ data to train our GPR model, which is still much less than what a neural network requires.

\subsection{Generalization to Different Containers}
\begin{figure}[t]
\centering
\includegraphics[width=0.35\textwidth]{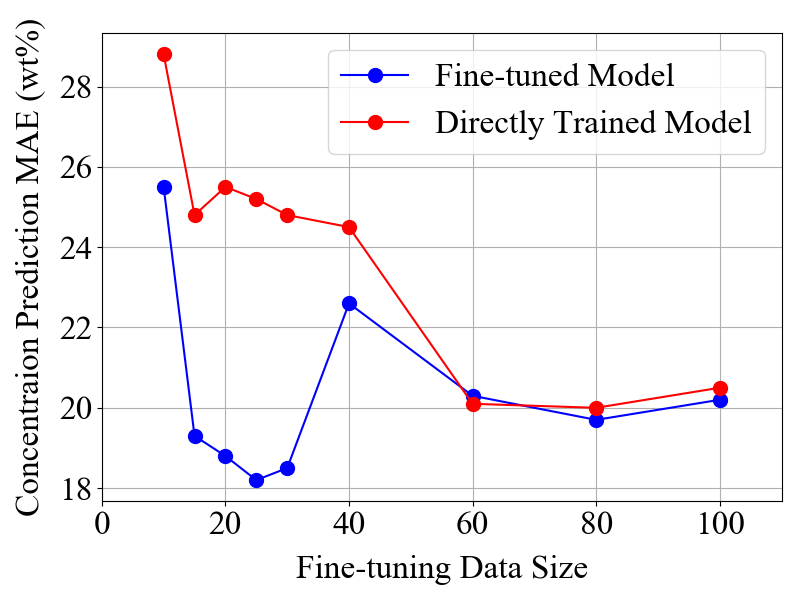}
\caption{The prediction error of sugar concentration on the cuboid container. Red is the model directly trained on the fine-tuning dataset. Blue is the model first trained on grooved cuboid container then fine-tuned on the fine-tuning dataset.}
\label{fig:exp6_tune_data_efficiency}
\end{figure}

Finally, we show that our model can generalize across similar-shaped containers with very little fine-tuning data. Given the GPR model trained in the grooved cuboid container (Fig. \ref{fig:exp2_gp_c_pred}), we fine-tune and test on the cuboid container using the approach in Section \ref{gcss}. The fine-tuning datasets have a size ranging from $10$ to $100$ data and are the same as the training datasets of the cuboid container used in Section \ref{data_efficiency}. We compared our results with the baseline model, which is a GPR model directly trained from the fine-tuning data. The concentration prediction MAE when using different amounts of fine-tuning data is shown in Fig. \ref{fig:exp6_tune_data_efficiency}. It takes $15$ datapoints to fine-tune our model. In fact, with more than $10$ fine-tuning datapoints, our fine-tuned model always has better or similar performance than the baseline model.

\section{Future Work}

\label{sec: future}
We believe our work is just the beginning of a sequence of super-human perception research based on dynamic tactile signals. One avenue to improve our approach on liquid properties estimation is making use of the dynamic tactile signals collected in the activation stage as well as the free-oscillation stage. In the future, we will survey models for signals in the activation stage and explore other ways to activate the system, such as perturbing in different directions. Other extensions of our approach include modeling the damped oscillation signal with more features and learning a more complex model than GPR models. We will also explore how the physics-inspired models can be applied to other dynamic tactile sensing tasks, such as estimating solid content properties.
\section{Conclusion}

\label{sec: conclusion}
In this paper, we present an approach to precisely estimate the liquid viscosity and volume in the container using dynamic tactile sensing. We introduce a physics-inspired model to interpret the damped oscillation signal after perturbation and relate the signal to the liquid properties. We then learn a GPR model to estimate liquid properties. In the experiment, our approach achieves 100\% classification accuracy among water, oil, and detergent. For the challenging sugar-water properties regression task, we can predict the height of the liquid with $0.56$ mm error and the sugar concentration with $15.3$ wt\% error. Our approach is also data-efficient, agnostic to volume, and can easily generalize to different types of liquid and containers. Our work elevates dynamic tactile sensing to a much higher precision level in this task and explains its principal mechanisms. We believe this lightweight learning approach based on simple physics can be applied to other dynamic tactile sensing tasks. 

\section*{Acknowledgments}
This work was supported by Toyota Research Institute. The authors would like to thank Yifan You for high-quality feedback in writing. We would also like to thank Uksang Yoo and Ruihan Gao for proofreading the paper.

\bibliographystyle{plainnat}
\bibliography{references}

\clearpage
\section*{Appendix}
\label{appendix}

\subsection{Experiment Results of Quadratic Regression Models}

\begin{figure}[!hbp]
  \centering
  \begin{minipage}[b]{0.48\linewidth}
    \subcaptionbox{\label{fig:exp2_quad_h_pred}}
      {\includegraphics[width=\textwidth]{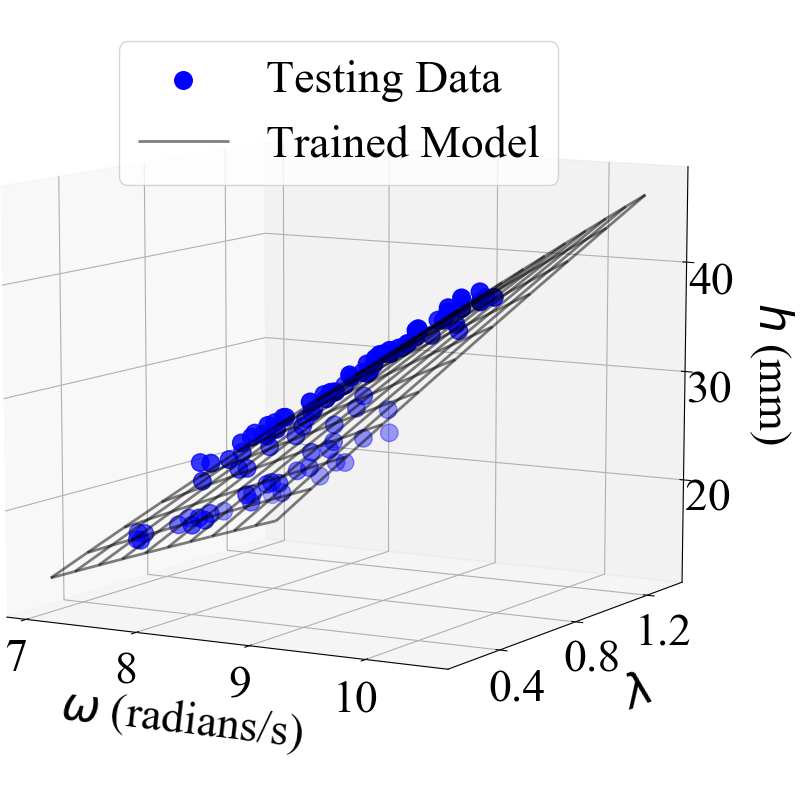}}
  \end{minipage}
  \hfill
  \begin{minipage}[b]{0.48\linewidth}
   \subcaptionbox{\label{fig:exp2_quad_c_pred}}
    {\includegraphics[width=\textwidth]{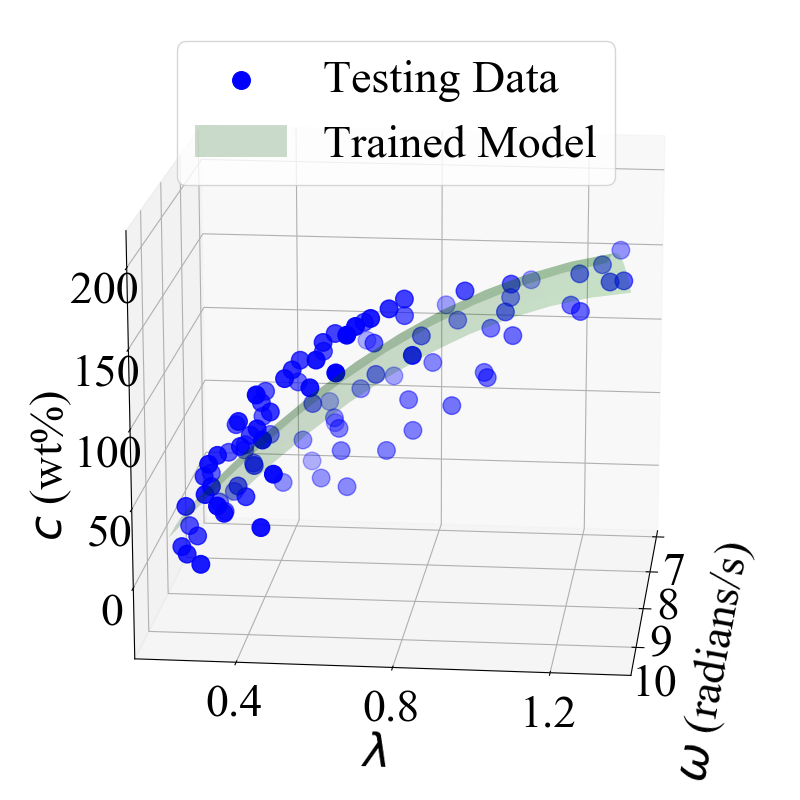}}
  \end{minipage}
  \hfill
  \begin{minipage}[b]{0.8\linewidth}
  \hfill
   \subcaptionbox{\label{fig:exp2_quad_h_conf}}
    {\includegraphics[width=0.48\textwidth]{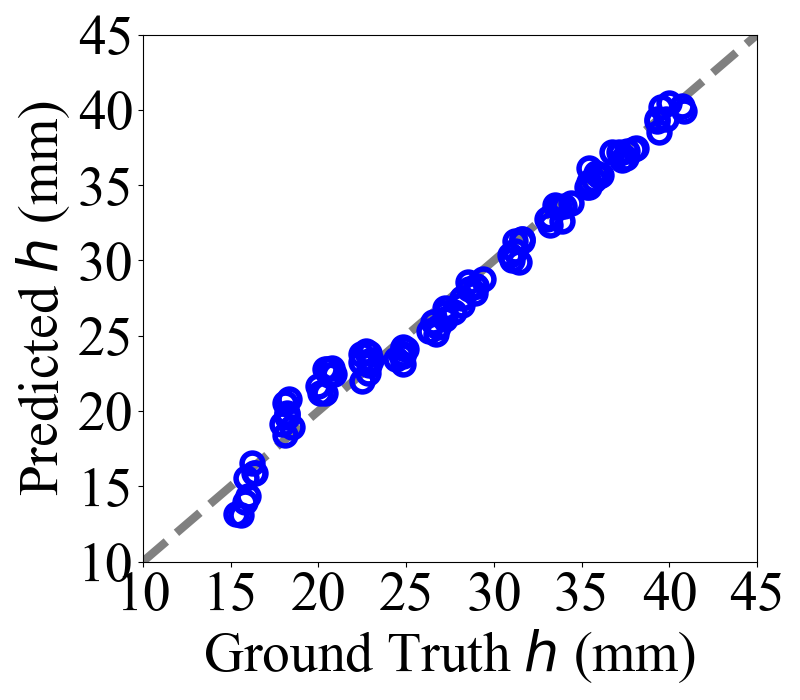}}
   \hfill
   \subcaptionbox{\label{fig:exp2_quad_c_conf}}
    {\includegraphics[width=0.48\textwidth]{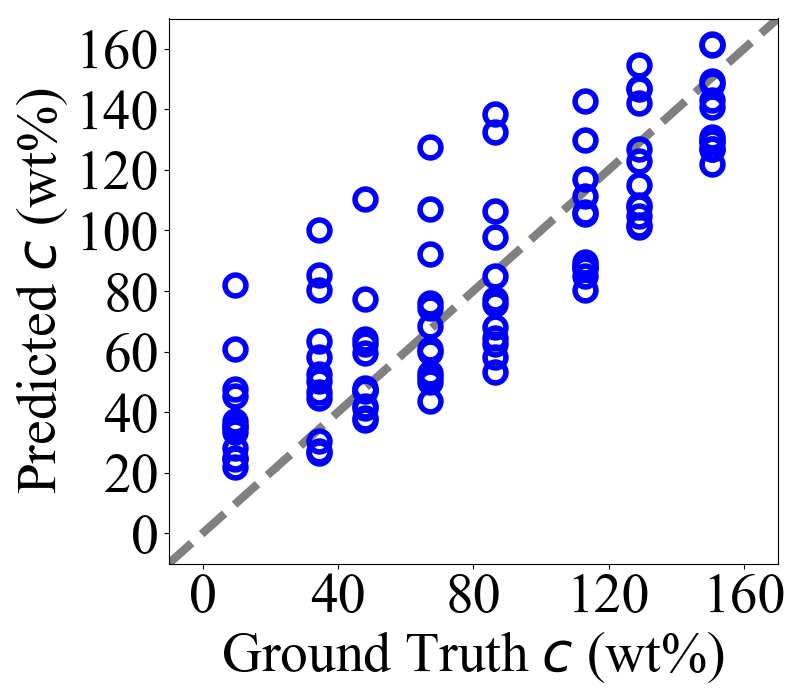}}
   \hfill
  \end{minipage}
  \caption{Prediction of liquid height and sugar concentration on the testing dataset using the quadratic regression model. (a) Relation between liquid height $h$ and decay rate $\lambda$, oscillation frequency $\omega$. (b) Relation between sugar concentration $c$ and decay rate $\lambda$, oscillation frequency $\omega$. (c) Prediction of liquid height. (d) Prediction of sugar concentration.}
  \label{fig:exp2_quad}
\end{figure}

In addition to estimating sugar-water concentration and liquid height using GPR models as shown in Section \ref{sugar_water}, we also train parameterized quadratic regression models to make the predictions as introduced in Section \ref{rlp}. The prediction result on the testing dataset is shown in Fig. \ref{fig:exp2_quad} and the training and testing errors are shown in Table \ref{tab:quad}. Quadratic models can estimate $h$, $c$, and $\mu$ with MAE of $0.84$ mm, $20.0$ wt\%, and $0.23$. The testing result using quadratic models is slightly worse than the non-parametric GPR models.

\begin{table}[h]
\centering
\begin{tabular}{|c|c|c|c|}  
\hline 
 & $h$ (mm) & $c$ (wt\%) & $\mu$ \\ 
\hline 
\hline 
Quadratic Model MAE (training dataset) & 1.13 & 21.1 & 0.23 \\ 
\hline 
Quadratic Model MAE (testing dataset) & 0.84 & 20.0 & 0.23 \\ 
\hline 
\end{tabular}
\caption{Prediction error of liquid height $h$, sugar concentration $c$, and $\log_{10}$ of viscosity $\mu$ using quadratic regression models.}
\label{tab:quad}
\end{table}

\end{document}